# Detection of Tomato Ripening Stages using Yolov3-tiny

Gerardo Antonio Alvarez Hernandez[1][0000−0002−4849−5319], Juan Carlos Olguin[1][0000−0002−2613−8345], Juan Irving Vasquez[1][0000−0001−8427−9333], Abril Valeria Uriarte[1][0000−0003−2222−303X], and Maria Claudia Villicaña Torres[3,2][000−0002−33−8416]

[1] Instituto Politécnico Nacional (IPN), Centro de Innovación y Desarrollo Tecnológico en Cómputo (CIDETEC), Ciudad de México, México
[2] Consejo Nacional de Ciencia y Tecnología (CONACYT), Ciudad de México, México
[3] Centro de Investigación en Alimentación y Desarrollo A.C (CIAD), Culiacán, Sinaloa

**Abstract.** One of the most important agricultural products in Mexico is the tomato (*Solanum lycopersicum*), which occupies the 4th place national most produced product . Therefore, it is necessary to improve its production, building automatic detection system that detect, classify an keep tacks of the fruits is one way to archieve it. So, in this paper, we address the design of a computer vision system to detect tomatoes at different ripening stages. To solve the problem, we use a neural network-based model for tomato classification and detection. Specifically, we use the YOLOv3-tiny model because it is one of the lightest current  deep neural networks. To train it, we perform two grid searches testing several combinations of hyperparameters. Our experiments showed an f1-score of 90.0% in the localization and classification of ripening stages in a custom dataset.

## 1 Introduction

One of the most important agricultural products in Mexico is the tomato (*Solanum lycopersicum*), which occupies the 4th place national most produced product. Likewise, it occupies the 9th place globally in the production of this product, with the amount of 3,370,877 tons in 2020 [1].

In the production of tomatoes, shape and color have a considerable influence on the expectations of their commercial value. In terms of shape, it is desirable to be round, globular, or oval, depending on the type and in terms of color, it has a uniform color ranging from orange to deep red, with no green shoulders. Its appearance should be smooth and with small scars corresponding to the floral tip and peduncle, in addition, it is firm to the touch, that is, it is not soft or easily deformed due to over maturity [2]. With this in mind, it is necessary to monitor the ripening stages of tomatoes during the production process, because whether it is not done correctly, it could have a poor quality tomatoes causing losses. This task is carried out by visual inspection with people trained to identify: the degree of ripening and fruit sizes, however, these methods are subjective.

Several methodologies have been proposed for the creation of systems that help in this task, such as Vishal et al. [3] where the authors propose a monitoring model using an Arduino, with color characteristics L*a*b to discern the ripening stage of the tomato. Anna et al. [4] propose instead of extracting the color, to use tomato florescence based on the reflection and emission of heat that the fruit has, with the drawback that the tomato must be in a certain temperature range to perform a correct classification. On the other hand, Kejin et al. [5] use these two characteristics, color and fluorescence using a colorimeter, with the drawback that the detection is only effective when the tomatoes are at a temperature of 25 degrees Celsius. As can be seen, these solutions focus on finding the decision pattern through the expert's knowledge , another way of solving is with the use of machine learning models so that the decision patterns are chosen automatically without the need of an expert such as the one of Aranda-Sanchez et al. [6] which uses the colorimeter to obtain the data and through the use of a Bayesian classifier determines the degree of ripeness of the tomato. Seeing that these systems focus only on the classification of tomatoes in their ripening stages, there are other systems that, in addition to classifying them, also locate them, commonly called detection systems, such as Nuttakarm et al.[7], the authors propose to use a R-CNN (region-based convolutional neural network) to locate the tomato, then the images are segmented using a k-means algorithm with 2 centroids $k = 2$ : area of interest (fruit) and uninterested region. The segmented areas of interest are converted to Hue-Saturation-Value (HSV) color space to extract characteristics, that are then used with a Support Vector Machine (SVM) to classify the ripening stage of the tomato. Dasom et al.[8] propose to use the R-CNN , but as a multiclass detector in which the classes are the different ripening stages, being able to detect the ripening stages with occlusions, but at the expense of a large computational cost when using this detection model.

In this work, we propose to solve the problem of tomato ripening stage detection to be used in an automatic tomato monitoring system. Therefore, we propose to solve this problem by means of a vision system implementing an existing detection model. This model is based on a one-step convolutional network with multi-class detection YOLOtiny-V3 [9]. This network was chosen because it is is one of the lightest current deep neural network architectures. We used a database of 3,000 images of tomatoes labeled with the class name, corresponding to their ripening stage, and a bounding box enclosing the tomato [19]. In addition, we set up an experimental design with various combinations of its hyperparameters to obtain 90.0% in f1-score.

The rest of the paper is organized as follows. Section 2 presents some related work on the detection and classification of tomato maturity stages. Section 3 describes the components that compose our proposed solution. Section 4 de-scribes the experiments performed, as well as the analysis of results, and Section 5 presents the conclusions and suggestions for future work.

## 2   Related Work

This section reviews the various research approaches that have currently been adopted to solve the problem of detecting and classifying ripening stages of tomato, as well as other fruits.

Starting with ripeness sorting, Dasom et al.[8] propose a model of tomato ripeness monitoring inside a hydroponic greenhouse. They use the fastest R-CNN for the process of detecting the tomato. Once the fruit region is detected, the k-means algorithm is used to separate the background from the fruit region, then it is converted from RGB to HSV color space extracting only the hue channel, to classify the ripeness using 6 color shades. On the other hand, Nashwa et al.[10] use Principal Component Analysis (PCA) to extract the characteristics of tomatoes in their different ripening stages in order to be delivered to a SVM to perform the classification, a Linear Discriminant Analysis (LDA) was also tested, but best classification score was obtained by the one-against-one multi-class SVMs system. Another work like the one by De Luna et al.[11] propose to locate and detect ripening stages of tomatoes, using two pre-trained convolutional models, the R-CNN and SDD (Single Shot Detector) with better performance of SDD to detect flower and fruit. For maturity classification, 3 basic classifiers were used: artificial neural network (ANN), k-nearest neighbors (K-NN) and SVM, the latter having the highest classification accuracy. Lui et al.[12] proposed an improved DenseNet model to detect tomato ripeness, to improve it, they proposed a structured sparse operation by splitting the convolution kernel into multiple groups, obtaining a computational reduction of 18% of the original network. It can detect tomatoes in a complex background and different sizes.

Rangarajan et al. [13] classify tomato leaf images with 6 different diseases. Using 2 different deep learning architectures AlexNet and VGG16net, a transfer learning approach, where Alexnet obtained the highest classification performance with a 97.99% accuracy. Hong et al. [14] test 5 deep networks Resnet50, Xception, MobileNet, ShuffleNet, Desnet121_Xception to detect tomato plant diseases, to decrease the computational time they used transfer learning and data augmentation. Desnet121_Xception obtained the best performance with 97.10% accuracy.

There are several approaches that have been used for fruit detection, such as Liu et al. [15] where the authors used a gradient-oriented descriptor (HOG) to train an SVM to detect tomatoes, followed by an FCR (false color removal) to eliminate false positives and an NMS (Non-Maximal Suppressor) to eliminate repetitions. Zhang et al. [16] implementing a convolutional network (CNN), they focused on testing different data augmentation methods. These were: rotation, scaling, Gaussian noise, salt noise, pepper noise, as well as combinations. In addition, the method uses t-Distributed Stochastic Neighborhood embedding (t-SNE) to remove poor-quality images to obtain a better dataset. Rotational enhancement, scaling and salt noise methods performed better, reaching an accuracy of 91.9%.

Y. Mu et al. [17] built a model that detects green tomatoes regardless of whether they are occluded or at different stages of ripening using an R-CNN coupled with Resnet 101 using COCO data for transfer learning. They obtained an average accuracy of 87.83% taking into account that the tomatoes were in their natural environment compared to others where they are in controlled conditions. Although with a high computational cost, Lui et al.[18] modified the YOLO v3 model for tomato recognition called YOLO-Tomato. it incorporates a DenseNet architecture to reuse image features, and replaces the traditional rectangular box with a circular one. They applied data augmentation with scaling and cropping operations in order to increase performance. This model is able to detect tomatoes, taken with various light intensities, achieving a prediction rate of 94.58%.

To summarize this review, our proposal detects the ripening stages using a single model, compared to the work of Dasom et al. [8] and De Luna et al. [11] whose approaches were to separate the detection by occupying two models, one for locating and one for classifying. Furthermore, the works of Nashwa et al. [10] and Rangarajan et al. [13] only classify ripening stages and do not detect them. And in the case of Luis et al. [[12],[18]], and Hong et al. [14] only detects only the tomato class in contrast to ours which solves a more particular problem which is to detect each of the tomato ripening stages.

## 3   Multi-class Detection of Ripening Stages

To achieve the detection of tomato ripeness, we propose a vision system that uses a general-purpose deep learning architecture to perform multi-class detection. This architecture is the YOLOv3-tiny which is a small version of Yolo v3 [9]. Given that our problem has not been reported previously, we train the network architecture using a custom dataset. To train the network, we design two grid searches, these searches allow us to find adequate network parameters.

We use a previous dataset [19], we labeled it and then we trained the network with various parameters. In the end, we validate it using detection thresholds to calculate accuracy, precision, recall, f1-score, and IOU.

### 3.1   Dataset

The dataset consists of 3,000 images of tomatoes, fruit of *Solanum Lycopersicum*, with a resolution of 800× 600 pixels, in JPG format and RGB channels. The original dataset was published in [19]. In that work, the dataset contains only the classes of each of the images (no bounding boxes). The tomato images were taken in controlled conditions, varying the position of the tomatoes to have a diversity of view of the object.

The 6 labels handled in the dataset are: Orange (500 of this class), Striped (500 of this class), Red (500 of this class), Red-Orange (500 of this class), Salmon (500 of this class) and Green (500 of this class ) (see Table 1).Therefore, the database is balanced.

For this work, we have manually labeled the bounding boxes enclosing the tomato found in each image, as well as the label indicating the ripening class of the tomato. This labeling was done by using the "labelImag" tool [20].

The images in the dataset are re-scaled to fit the input size required by Yolov3-tiny, which is 416 x 416 pixels.

| Image | Label | Image | Label | Image | Label |
|---|---|---|---|---|---|
| 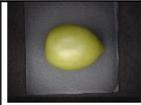 | Green | 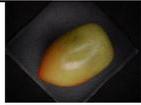 | Striped | 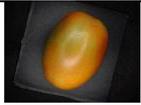 | Salmon |
| 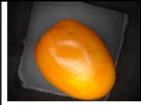 | Orange | 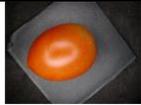 | Red-orange | 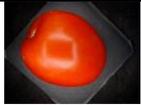 | Red |

**Table 1.** Examples of dataset

### 3.2  Deep learning architecture YOLOv3-tiny

The deep learning algorithm used belongs to the You Only Look Once (YOLO) family. This family of algorithms exploits the use of convolutional neural networks for object detection. They are one of the fastest object detection algorithms available and are a good choice for real-time detection without compromising accuracy. YOLO is an object detection architecture proposed by Joseph Redmon et al. in 2015 [21] and improved with the second version in 2016 (YOLOv2) [22] and with it was third version in 2018 (YOLOv3) [23].

The YOLOv3 structure consists of six types of layers, defined as net, convolutional, max-pooling, yolo, route, and upsample. The net layer configures the parameters of the entire network. The convolution layer has three modules: convolutional operation, batch normalization, and activation function. In addition, the convolution operation includes feature map conversion and general matrix multiplication.

Feature learning is done through convolutional layers, which have a similar structure to ResNet, in YOLOv3 they are called "Residual Blocks" and are used for feature learning. Residual blocks consist of several convolutional layers and jump connections. One feature of YOLOv3 is that it performs detection at three different scales. The scales are 13×13, 26 ×26, and 52 × 52. The depth of model is calculated with the following operation:

$$depth_{yolo} = (5 + n_{clases}) \times n_{anchors} \qquad (1)$$

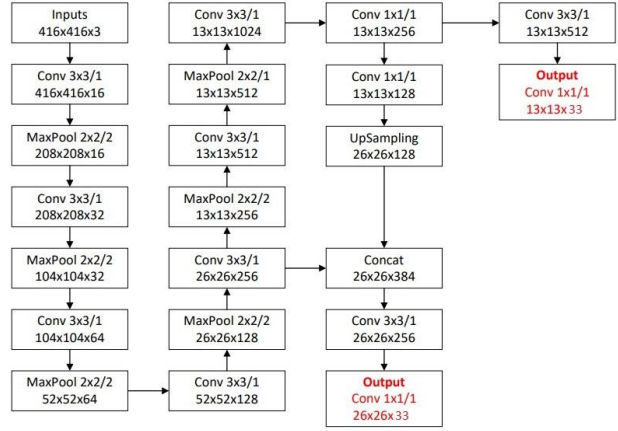

**Fig. 1.** Block structure of the YOLOv3-tiny architecture used for the detection of this case study.

$$depth_{yolo} = (5 + 6) \times 3 \qquad (2)$$

$$depth_{yolo} = 33 \qquad (3)$$

In this case, there are 6 maturity classes and 3 anchors for each scale, which results in a depth of 33 in each yolo output. The YOLOv3 architecture, considering as input images the images of this work, whose dimension is $416 \times 416 \times 3$ (where the three represents the RGB channels).

There is another version of YOLO, which has a reduced depth of the convolutional layers and it is called YOLO v3-Tiny. It was also proposed by Pranav Adarsh [9]. A characteristic of this architecture is that it has only convolutional layers for feature extraction, which causes that the execution speed to increase significantly. YOLO v3-Tiny uses a max-pooling layer and thus reduces the image in the convolution layer. It predicts using a three-dimensional tensor containing objectivity score, bounding box, and class predictions on two different scales. It divides an image into $S \times S$ grid cells. For final detections, ignore the bounding boxes for which the objectivity score is not the best using the non-maximum suppression method (NMS).

The bounding box prediction occurs at two feature map scales, which are $13 \times 13$ and $26 \times 26$. The YOLOv3-tiny architecture for input images with dimension $416 \times 416 \times 3$ is shown in Fig. 1.

## 4   Experiments

In this work, two different grid searches were performed to find the most promising result, each of these grids will be detailed in this section, as well as the results obtained and their analysis. The performance of each experiment was evaluated

| ID | Factor 1 | Factor 2 |
|----|----------|----------|
| 1  | $1 \times 10^{-3}$ | Adam |
| 2  | $1 \times 10^{-3}$ | SGD |
| 3  | $1 \times 10^{-5}$ | Adam |
| 4  | $1 \times 10^{-5}$ | SGD |

**Table 2.** Exploratory grid search

using the following performance measures: accuracy, precision, recall, f1-score [24] and IOU [25].

### 4.1 Design of Experiment

Two grid searches were used, one is the exploratory grid search where several factors were tested to see their behavior and to investigate in the search for the most suitable ones. With this, it was observed which factors were the best performers. From there, another grid search was created to make a more specific search with the factors chosen by the exploratory grid search. Images were divided into 2 batches 2,400 images for training and 600 images for validattion. The training was performed using Google Colab platform with Python 3, Keras library version 2.1 and Tensorflow library version 1.15.

To evaluate the classification performance, two parameters were used as thresholds, a score of 0.7 and IOU of 0.5, that will be used to filter the pure results of the model, those that did not meet the results provided by the model were not considered as detected. The accuracy, precision, recall, and f1-score metrics were calculated using a function made by us in python. In the calculation of the metrics the total number of labels was considered as the number of labels that were recognized of the images in which a tomato was detected and not the total number of images in the test set.

### 4.2 Exploratory Grid Search

For the exploratory grid search, two hyperparameters were used as factors: the optimizer and the learning rate. The batch size of 64 and the number of epochs of 100 were kept as orthogonal factors. Since the computational cost was the reason for choosing 100 epochs to keep the training time within a reasonable time margin. The grid search is shown in table 2.

The training behavior of this exploratory grid search can be seen in Figure 2. It is observed that the loss curve when using the $1 \times 10^{-3}$ learning rate, regardless of which optimizer is used, decays faster compared to the $1 \times 10^{-5}$ rate.

With this first experiment (see table 3) the best performing was the ADAM optimizer with the learning rate of $1 \times 10^{-3}$ obtaining a score of 0.93 in the f1-score . On the other hand, where the learning rate was $1 \times 10^{-5}$ it was not possible be calculated, since the found weights were not enough to reach the thresholds established for performance measurement. The conclusion of this grid is that it

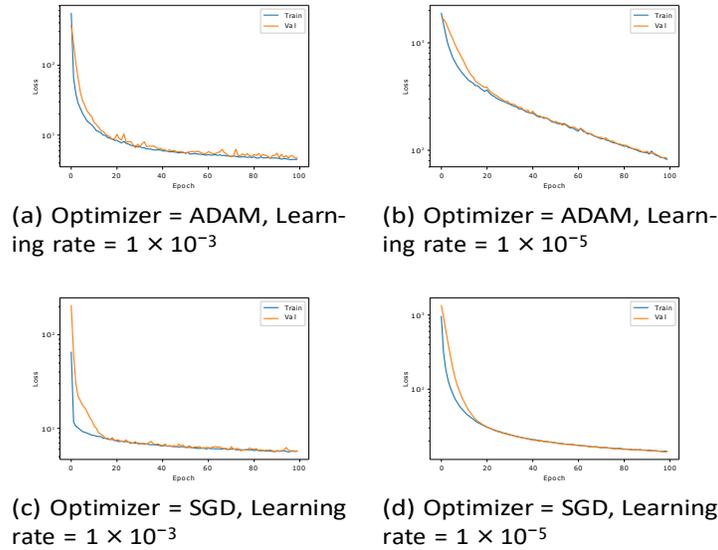

(a) Optimizer = ADAM, Learning rate = $1 \times 10^{-3}$

(b) Optimizer = ADAM, Learning rate = $1 \times 10^{-5}$

(c) Optimizer = SGD, Learning rate = $1 \times 10^{-3}$

(d) Optimizer = SGD, Learning rate = $1 \times 10^{-5}$

**Fig. 2.** Graphics of training performance on the exploratory grid search.

| Experimets | | Performance thresholds | | Orthogonal Parameters | | Metrics | | | | | Label |
|---|---|---|---|---|---|---|---|---|---|---|---|
| Optimizer | Learning rate | Score | IOU | Epochs | Batch size | Accuray | Precision | Recall | F1-score | Average IOU | Not detected |
| ADAM | $1 \times 10^{-3}$ | 0.7 | 0.5 | 100 | 64 | **0.95** | **0.93** | **0.95** | **0.93** | **0.86** | **117** |
| ADAM | $1 \times 10^{-5}$ | 0.7 | 0.5 | 100 | 64 | - | - | - | - | - | 600 |
| SGD | $1 \times 10^{-3}$ | 0.7 | 0.5 | 100 | 64 | 0.91 | 0.71 | 0.89 | 0.74 | 0.76 | 531 |
| SGD | $1 \times 10^{-5}$ | 0.7 | 0.5 | 100 | 64 | - | - | - | - | - | 600 |

**Table 3.** Table of performance metrics obtained by the exploratory grid search.

is convenient to use a learning rate of $1 \times 10^{-3}$ as it was the one with the best results. In the case of the ADAM optimizer, it is convenient to use it, since it was the one that had the best performance. With this, we perform an exploit grid search now keeping the Adam optimizer fixed using a series of learning rate values.

### 4.3   Exploitation Grid Search

For the exploitation grid search, only the learning rate was considered as a factor. The batch size of 64 and the number of epochs of 200 were kept as orthogonal factors. As in this grid, the number of experiments is smaller, therefore, it was decided to increase the number of epochs to 200 so that the training takes enough time to converge. And from the conclusions of the exploratory grid search, the ADAM optimizer will be kept fixed for being the one with the best performance. The grid search is shown in the table 4.

Detection of Tomato Ripening Stages using Yolov3-tiny     9

| ID | Factor 1 |
|----|----------|
| 1  | $1 \times 10^{-3}$ |
| 2  | $1 \times 10^{-4}$ |
| 3  | $1 \times 10^{-5}$ |

**Table 4.** Exploitation grid search

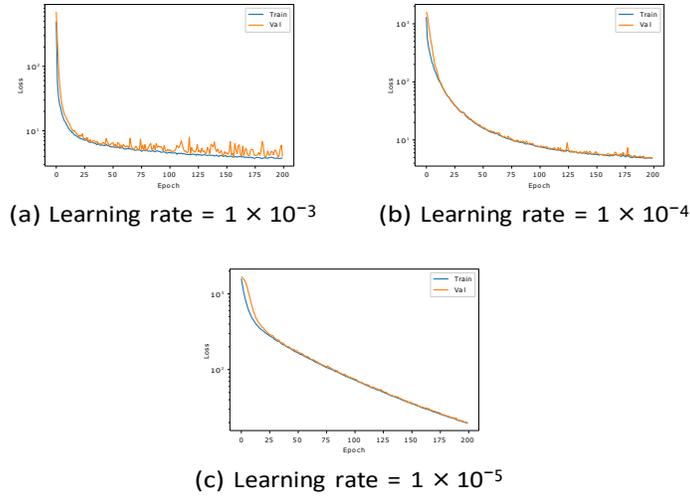

(a) Learning rate = $1 \times 10^{-3}$     (b) Learning rate = $1 \times 10^{-4}$

(c) Learning rate = $1 \times 10^{-5}$

**Fig. 3.** Graphics of training performance on the exploitation grid search.

The training behavior of this exploitation grid search can be seen in Figure 3. The table of metrics obtained from the exploitation grid search is shown in table 5.

### 4.4 Analysis of Results

The results obtained in the grid search of exploitation (see table 5) the best metrics was the one with the learning rate of $1 \times 10^{-4}$, but with less amount of detected images. While, the $1 \times 10^{-3}$ rate had slightly lower performance, it achieved the highest number of detected images. In addition, the average IOU in both models are identical, which indicate a similar behavior on average over all classes. In the case of the rate of $1 \times 10^{-5}$ due to the low number of images detected, we did not obtain its metrics. For a deeper analysis of the behavior of these two experiments, we present their confusion matrices (see figure 4). In these matrices, the labels were coded to numbers for better visualization.

| Experimets | Performance thresholds | | Orthogonal Parameter | | | Metrics | | | | | Label |
|---|---|---|---|---|---|---|---|---|---|---|---|
| Learning rate | Score | IOU | Optimizer | Epochs | Batch size | Accuray | Precision | Recall | F1-score | Average IOU | Not detected |
| $1 \times 10^{-3}$ | 0.7 | 0.5 | ADAM | 200 | 64 | 0.91 | 0.92 | 0.90 | 0.90 | **0.87** | **84** |
| $1 \times 10^{-4}$ | 0.7 | 0.5 | ADAM | 200 | 64 | **0.93** | **0.94** | **0.92** | **0.92** | 0.87 | 137 |
| $1 \times 10^{-5}$ | 0.7 | 0.5 | ADAM | 200 | 64 | - | - | - | - | - | 531 |

**Table 5.** Table of performance metrics obtained by the exploitation grid search

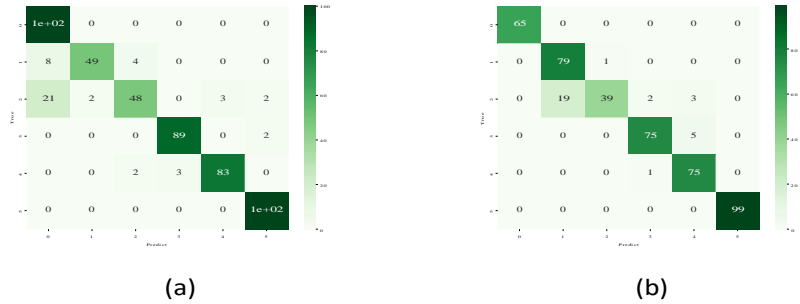

(a)                                          (b)

**Fig. 4.** Confusion matrix of the two experiments, (a) confusion matrix for $1 \times 10^{-3}$ experiment and (b) confusion matrix for $1 \times 10^{-4}$ experiment. The number 0 represents the red class, 1 the red-orange, 2 orange, 3 striped, 4 salmon and 5 green.

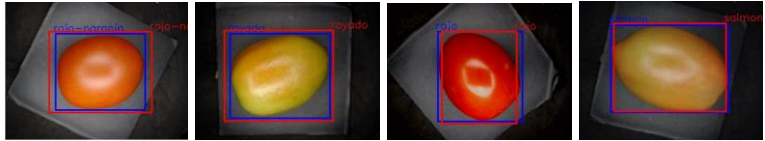

**Fig. 5.** Detection result with parameters: ADAM optimizer and learning rate $1 \times 10^{-3}$

With these matrices it is observed that the experiment with learning rate of $1 \times 10^{-4}$ detects well the classes 0 and 5, on the other hand, the one with rate of $1 \times 10^{-3}$ detects more images of that class but performs more classification errors. The experiment with rate of $1 \times 10^{-3}$ detects more images of class 3 and 4 compared to the other one, but this one presents more errors when classifying these classes. Class 2 in both experiment had the highest error rate. So for us, the best experiment was the $1 \times 10^{-5}$ rate experiment because it detects more images. Figure 5 shows an example of the detection results obtained by the best experiment.

## 5   Conclusion and Future Work

A tomato ripening stage detection system was constructed, obtaining an accuracy of 91.0% at the best score. The detection performance of the ripening stages was 90.0% in the f1 score. The relevance of our configuration is that the location of the tomatoes is very close to true, this is observed in the average IOU which means that when the tomato is detected the model is 87% sure of its location in the image. The point of improvement for this model is to make it capable of detecting tomatoes in their natural habitat so that it can be used by a harvesting robot, as it is currently only capable of detecting them under very controlled conditions.

In future work, a data augmentation optimization stage will be introduced to the system so that the model will be able to detect tomatoes in varying circumstances.